# DEEP TRAJECTORY FOR RECOGNITION OF HUMAN BEHAVIOURS

Tauseef Ali and Eissa Jaber Alreshidi

*ABSTRACT*

*Identifying human actions in complex scenes is widely considered as a challenging research problem due to the unpredictable behaviors and variation of appearances and postures. For extracting variations in motion and postures, trajectories provide meaningful way. However, simple trajectories are normally represented by vector of spatial coordinates. In order to identify human actions, we must exploit structural relationship between different trajectories. In this paper, we propose a method that divides the video into N number of segments and then for each segment we extract trajectories. We then compute trajectory descriptor for each segment which capture the structural relationship among different trajectories in the video segment. For trajectory descriptor, we project all extracted trajectories on the canvas. This will result in texture image which can store the relative motion and structural relationship among the trajectories. We then train Convolution Neural Network (CNN) to capture and learn the representation from dense trajectories. . Experimental results shows that our proposed method out performs state of the art methods by 90.01% on benchmark data set.*

## 1. INTRODUCTION

Automatic recognition of human actions play the importance role and is an active topic in computer vision and pattern recognition research [1], [2], [3]. Action recognition has wide range applications in area of automated surveillance [7], [8], [9] [10], in-home elder monitoring [4], [5], [6], and public security [11], [12], [13] and customer behavior [14], [15], [16]. Technically speaking, the task of human action is to identify and recognize actions and behaviors of one or more persons from video sequences. For example, an in-home elder is doing an exercise or doing some other activity by using his/her hands, arms, legs, and other body parts. There are two ways to observe his/her action, 1) bare eyes, 2) camera. We can easily understand and classify his/her action with the bare eyes, since our minds are trained for specific categories. For example, we can easily classify the behavior of a person is walking or running. Such manual classification is tedious and hectic job and we need automatic methods that automatically can identify and classify these actions. For example, in rehabilitation process, it is important to observe and monitor human actions for a long period of time [17], [18]. It is impossible for humans to manually analyze these action for long period of time due to limited capabilities [19], [20]. Therefore, we need some method that can automate this process.

The alternative solution is to develop a virtual machine [21], [22], [23] that can accurately analyze and understand humans' actions from the videos in order reduce to human labor and error. For example, in smart rehabilitation, a virtual analyst with the help of camera can automatically understand and analyze his/her behavior. There are many applications of such virtual analyst. We can prevent the patient from injuries by predicting his/her behavior. Other important applications include video surveillance, entertainment. Motivated by the importance of automatic action recognition methods, several work reported in literature [24], [25], [26], [27] with aim to automatically classify human actions in analyzing a large-scale of video data and provide understanding on the current state and future state.

Cutler et al. [30] [31] detect and identify the periodic motion in extremely low resolution videos. They compute similarity among the different frames of the video sequence and find whether an action is periodic. The limitation of this method is that they only use appearance features while we argue that similarity among the appearance features cannot distinguish the variation of posture. In order to overcome the above limitation, motion gradient information was used to classify actions. Efros et al. [32] introduced a novel motion descriptor which was based on the optical flow and motion similarity. They compute optical flow from consecutive frames and smooth out the optical flow in four separated channels. Spatial-temporal motion descriptors is then computed to identify different actions. Chaudhry et al. [36] proposed a histogram of oriented Optical flow (HOOF) and used Binet-Cauchy kernels to classify human actions. HOOF was very effective in identifying the human action since this method can alleviate the effect of noise, scale and motion variation. From the above discussion, it is revealed that representation of motion information play an important role in identifying human actions. In the most recent literature, the fisher vector representation integrated with improved dense trajectory descriptor is most effective way of capturing motion information. Originally, dense trajectory features are computed by sampling and tracking dense points from the each frame in multiple scales, while the improved trajectory feature is computed by estimating the camera motion and replaces the bag-of-features with Fisher vector. With such adaptation, they achieved state of the art performance on several action recognition datasets. However, in both these approaches, trajectories are represented as vector of spatial coordinates, hence losing structural relationship between different trajectories.

We propose a method to encode the trajectories in more effective way. Motivated by the great success of CNNs, we propose a novel trajectory descriptor based on CNN. Our proposed framework has the following steps: 1) Extract dense trajectories from multiple consecutive frames. 2) Project extracted trajectories on the canvas. This will result in a texture image. 3) Base on texture image, we utilize and train a convolutional neural network to learn compact representation of dense trajectories. From the experiments, we quantitatively show that our propose framework outperforms other state-of-the-art methods

## 2. PROPOSED METHODOLOGY

The proposed methodology starts by dividing the video into N number of temporal segments. We then compute dense optical flow between multiple consecutive frames using method in [41]. The reason of computing dense optical flow to extract motion information. We then extract trajectories by particle advection approach. After extracting trajectories, we then project those on the canvas resulting in a texture image. We then utilize and train CNN to learn structural relationship from texture images.

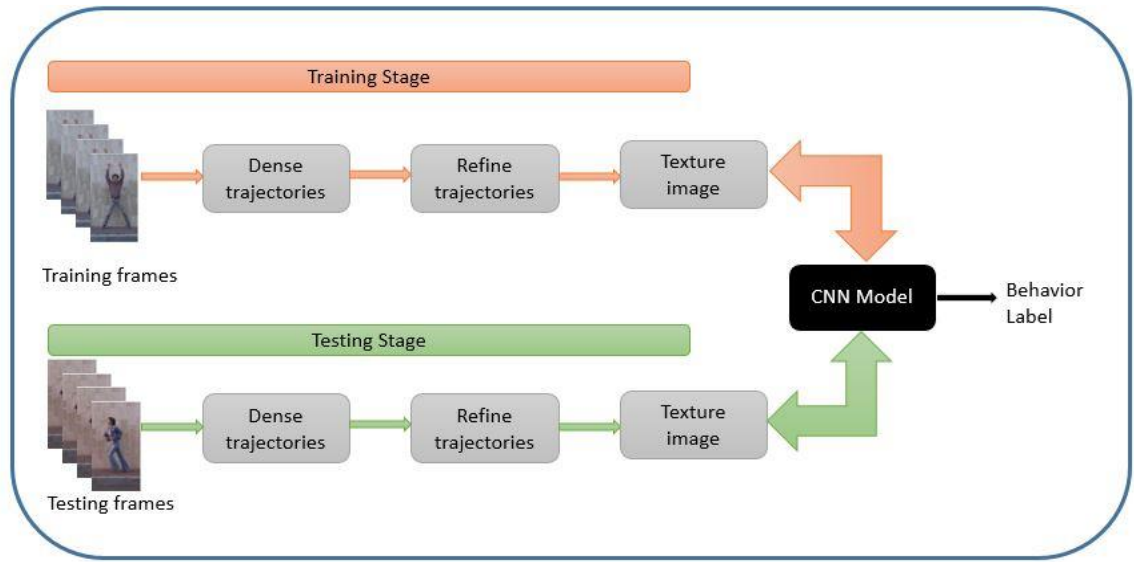

Figure 1: Pipeline of our proposed framework

## 2.1 OPTICAL FLOW COMPUTATION

The primary step to extract motion information is to compute dense optical flow between multiple consecutive frames. For computing high accurate optical flow we employ methods [42][60] where gray value consistency, gradient constancy and smoothness in multi-scale constraints. Consider a point i in the image at time t of a segment: its flow vector $Z_{i,t} = (X_{i,t}, V_{i,t})$ includes the location of point and its velocity is represented by $V_x$ and $V_y$. Where $V_x$ and $V_y$ represents the gradients in horizontal and vertical directions. After computing optical flow for each pixel, we have now motion field where higher magnitudes corresponds to the foreground and lower magnitude pixels represents the pixels belong to the background.

## 2.2 PARTICLE ADVECTION

After computing optical flow for whole video segment, the next step is to generate dense and long trajectories [43][44][59]. We overlay grid of particles over the first optical flow field. In first optical flow field, initial location of each particle is called the source location of the particle. In order to capture better representation of motion, we keep the size of the grid as same as resolution of the frame. The size of particle is same as size of the pixel. However, this arrangement will incur high computational costs. In order to reduce the computation cost and extract better representation of motion, we slightly reduce the resolution of the grid. During the process of advection, we maintain two separate flow maps, one to maintain the horizontal coordinates and other to maintain vertical components. These map in general store the initial and following positions of the point trajectories evolved during advection process.
The trajectories obtained through this process are applicable to structured crowds but for the unstructured crowds [45][46][58], where people have un-predicted behaviors, such trajectories do capture the actual motion information and flows. The reason is that in unstructured crowds, the people move in arbitrary directions and in most of the case there is chance that particle belong to a person will lose its path and become the part of different motion pattern. In this case, the trajectory becomes unreliable and erroneous and is not suitable to be considered. In order to tackle the above problem, we modify the above equation in the following way:

$$X_{(i,t+1)} = X_{(i,t)} + F(X_{(i,t)}) * B_i$$

Trajectories obtained using the above equation stop the trajectories drifting from one motion flow to other motion flow. Trajectories obtained by this method precisely capture the motion information in any complex situation.

After trajectories obtained using the above process, some of the trajectories will belong to the foreground while other trajectories will belong to background and noise [47][57], [48], [49]. In order to make the process efficient and effective, we remove trajectories belong to the background and noisy. To this end, we compute spatial extent of each trajectory by calculating the Euclidean distance between the start and last points of the trajectories. From our experiments, we observed that trajectories belong to noise and background are generally smaller in length. We use this information and set a threshold value on the length of trajectories. We remove those trajectories whose length is less than the specified threshold.

### 2.3 GENERATING TEXTURE IMAGE

After extracting trajectories, the next step is to generate texture image by projecting trajectories on the canvas. It is worthy to mention that most of the state-of-the-art techniques treat video in 3D space and extract features from spatial and temporal dimension. This method is very effective for small videos but incur high computation cost when large volume of video data is required to be processed. In order to address this problem, we propose a novel way to covert video into a two-dimension space so that can handled and processed easily.

Let Trajectory $T$ is represented by $(x, y, u, v)^t$, where $t$ is the frame number, x and y are the horizontal and vertical coordinates whereas u and v represents the velocity vectors. We then simply project trajectories from the multiple frames onto canvas C according to the following formulation.

$$C_{s,t} = \sqrt{u^2 + v^2}$$

Where $s$ is the canvas index at time $t$.

### 2.4 TRAJECTORY DESCRIPTOR BASED ON CONVOLUTION NEURAL NETWORK

In order to learn more compact representation of trajectories, CNN is employed to process the texture image. The main advantage of training CNN with texture image is that it learns structural relationship among trajectories. All texture images extracted from the video segments are stacked together and used as one multi-channel input for

CNN. Our CNN model is shown in Figure 1. It consists of four convolution layers. Each convolution layers is followed by ReLU layer. Convolution layers are then followed by max-pooling layers with two normalization layers and three fully connected layers. The main function of the convolution layer is to perform convolution operation on the feature map of the previous layers and extract features from the local neighborhood. Then additive bias is applied and the resulting feature map is passed to ReLU activation function. Generally, the value of a unit at position (x, y) of the $j^{th}$ feature map in the $i^{th}$ layer is given by $V^{xy}$ using the following formulation.

$$v^{xy} = a\left(b_{ij} + \sum_{m=0}^{M-1} \sum_{w=-\Delta}^{\Delta} \sum_{h=-\Theta}^{\Theta} \omega_{ijm}^{wh} v_{(i-1)m}^{(x+w)(y+h)}\right)$$

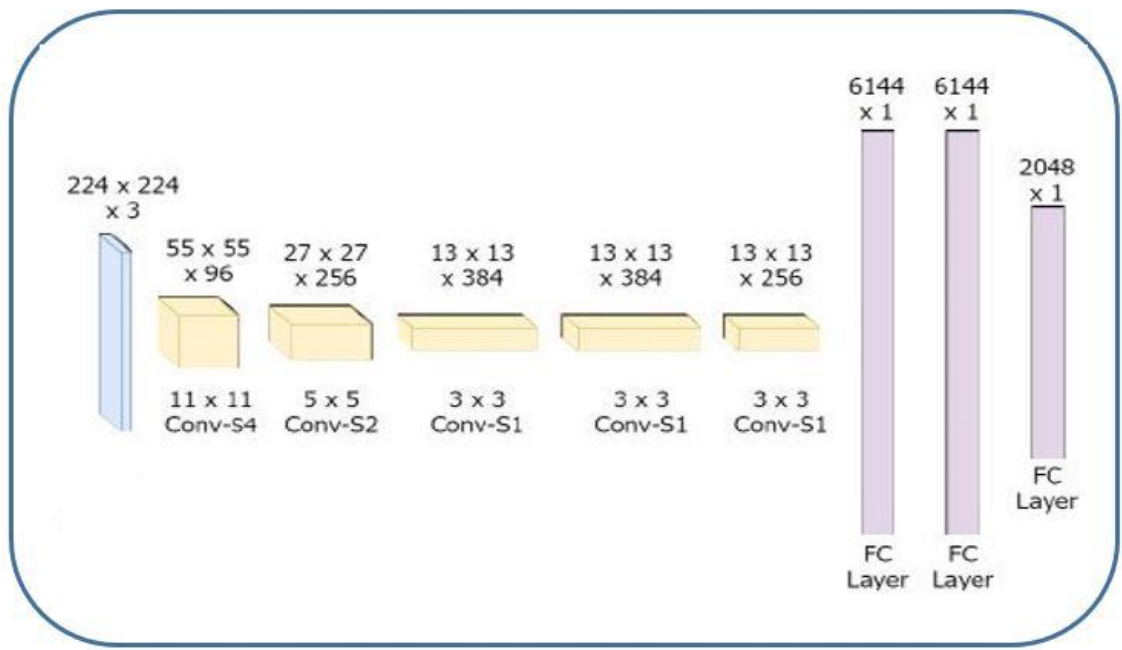

Figure 2: Convolution Neural network for human behaviour recognition

Where value of ▲ is half of width and θ is the half of height of an input image. $a$ (.) is the activation function, $b_{ij}$ is the bias of feature map. During training phase, we use gradient descent method.

The pooling layers pool over the local neighbourhood of the feature map in the previous layer result in reducing the resolution of feature map. In this way, pooling layers enhance the invariance to distortions on the input feature map.

We first train CNN on the training set. After training phase, we re-run the CNN model on the both training and testing sets and ignore the back propagation computation. We store all the parameter values from the fourth convolution layer at every iteration as the final representation of the video segment.

## 3 EXPERIMENT RESULTS

We use the publicly available benchmark dataset [54] for performance evaluation and comparison. The dataset contains 93 video sequences belong to 10 human actions. The actions are: bend, jump, and jack, jump forward- on-two-legs, jump-in-place-on-two-legs, run, gallop sideways, skip, walk, wave-two-hands, and wave one-hand). These actions are performed by 9 different actors. All video sequences in the dataset have the resolution of 180x144 pixels and the length of each video sequence is four seconds with average of 50 fps. The dataset also include extracted foreground, obtained by background subtraction.

In our experiment, we first compute optical flow for only the foreground objects excluding the background in order to reduce the computation time. In our experiments we use fixed configuration of CNN. The size of texture image is fixed to 165 x 165 pixels.

For training, we use 2/3 of the dataset and the rest for testing. In our experiments, in order to rigorously evaluate the performance of our proposed method, we randomly select six sequences in each action as a training set and the rest for testing.

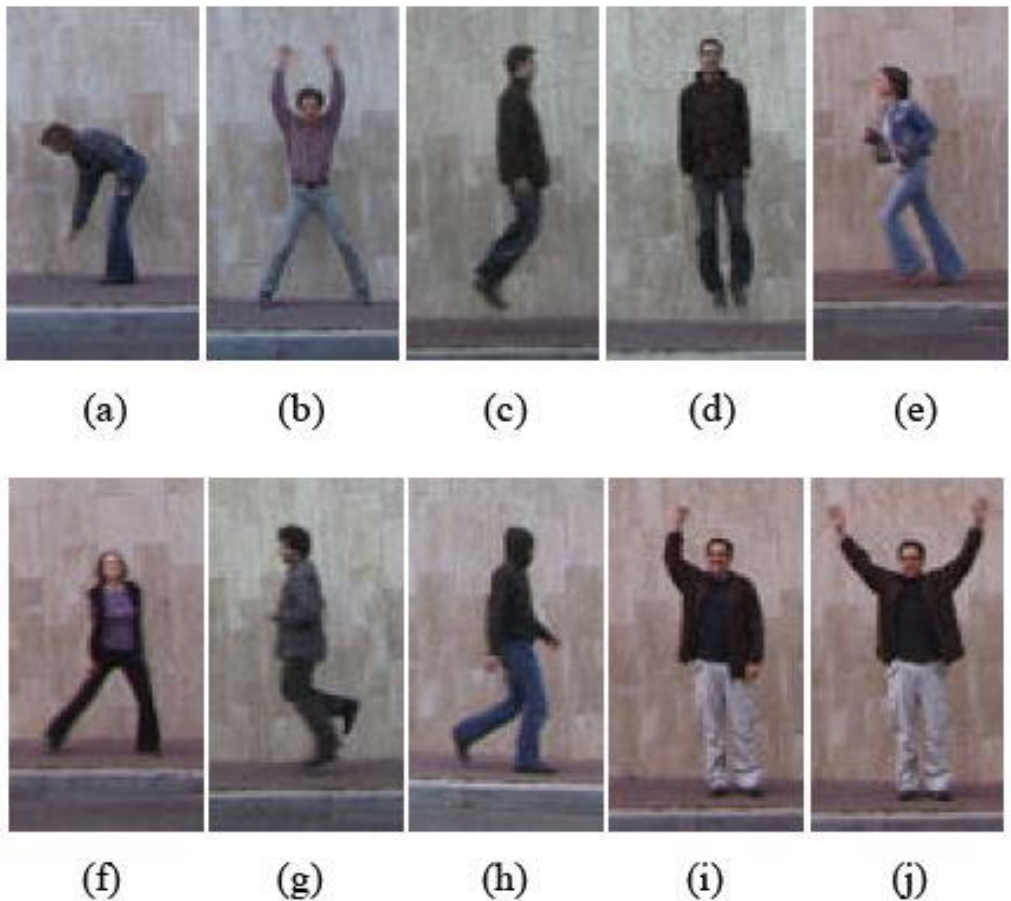

**Figure 2.** Sample frames from the dataset contain that 10 actions of 9 persons consists of (a) bend, (b) jack, (c) jump, (d) jump, (e) run, f) side, (g) skip, (h) walk, (i) wave1, (j) wave2

Figure 3, shows that confusion matrix for classification, where we train the classifier on one behavior and test it on the other behaviors. The average classification rate is

89.17%. However, there are some false positives due to the misclassification of bend with pjump. It attributes to the fact similar motion vectors are generated due to the person stand still before and after bending down. In skipping, 3/4 sequences are classified as running. It was expected because their motion and posture are very likely to each other. In addition, the misclassification of pjump with jump and wave2 with wave1 cause by similar pose too.

|       | Bend | Jack | Jump | Pjump | Run  | Side | Skip | Walk | Wave1 | Wave2 |
|-------|------|------|------|-------|------|------|------|------|-------|-------|
| Bend  | 0.87 | 0.00 | 0.00 | 0.33  | 0.00 | 0.00 | 0.00 | 0.00 | 0.00  | 0.00  |
| Jack  | 0.00 | 1.00 | 0.00 | 0.00  | 0.00 | 0.00 | 0.00 | 0.00 | 0.00  | 0.00  |
| Jump  | 0.00 | 0.00 | 1.00 | 0.00  | 0.00 | 0.00 | 0.00 | 0.00 | 0.00  | 0.00  |
| Pjump | 0.00 | 0.00 | 0.33 | 0.67  | 0.00 | 0.00 | 0.00 | 0.00 | 0.00  | 0.00  |
| Run   | 0.00 | 0.00 | 0.00 | 0.00  | 1.00 | 0.00 | 0.00 | 0.00 | 0.00  | 0.00  |
| Side  | 0.00 | 0.00 | 0.00 | 0.00  | 0.00 | 1.00 | 0.00 | 0.00 | 0.00  | 0.00  |
| Skip  | 0.00 | 0.00 | 0.00 | 0.00  | 0.75 | 0.00 | 0.74 | 0.00 | 0.00  | 0.00  |
| Walk  | 0.00 | 0.00 | 0.00 | 0.00  | 0.00 | 0.00 | 0.00 | 1.00 | 0.00  | 0.00  |
| Wave1 | 0.00 | 0.00 | 0.00 | 0.00  | 0.00 | 0.00 | 0.00 | 0.00 | 1.00  | 0.00  |
| Wave2 | 0.00 | 0.00 | 0.00 | 0.00  | 0.00 | 0.00 | 0.00 | 0.00 | 0.67  | 0.33  |

For quantitative analysis and comparisons, we also compare our method with other reference methods and the results are reported in Table 1. The conventional classification methods use the strategy of leave-one out with nearest neighbor while we used hold out method. In our method, the sequences in training set are not used in the testing process. For Leave-one-out with X samples, the model is train on all data except for one sample and test the model with the sample in each time X. We then compute the average error of X time. So every data used in testing once and in training X − 1 times. From the Table 1, it is obvious that our method produces lower error rate compare to other state-of-the-art methods.

| METHODS          | ERROR RATE |
|------------------|------------|
| Khan et al [24]  | **89.17**  |
| Ullah et al [43] | **95.24**  |
| Kong et al [3]   | **85.69**  |
| Proposed         | **72.11**  |

## 4 CONCLUSION

In this paper, we proposed a deep model approach for recognizing human behaviors. We proposed a novel approach to exploit motion information by projecting trajectories on the canvas. In this way, we generate texture image that captures the structural relationship among different trajectories. We then learn deep trajectory descriptor based on CNN. We demonstrated the capability of our approach in capturing the dynamics of different classes by extracting these features. The main advantage of the proposed method is its simplicity and robustness.